\documentclass[letterpaper]{article} 
\usepackage{times}  
\usepackage[hyphens]{url}  
\usepackage{graphicx} 
\usepackage{graphicx}  
\usepackage{geometry} 
\usepackage{amsfonts,amsmath,amssymb,amsthm} 
\usepackage{booktabs} 
\usepackage{makecell}
\usepackage{listings}

\newcommand{\citeauthor}{\cite}
\newcommand{\citeyear}{\cite}

\newcommand{\liff}{\leftrightarrow}

\newcommand{\eq}{\!=\!}

\newcommand{\lleq}{\!\leq\!}
\newcommand{\lland}{\!\land\!}

\newcommand{\init}{_\text{\emph{init}}}

\newcommand{\Eff}{E\!f\!f}

\newcommand{\ttt}[1]{\texttt{#1}}
\newcommand{\der}[2]{{\scriptstyle\frac{\mathrm{d}}{\mathrm{d#2}}}(#1)}
\newcommand{\lab}[1]{\noindent\textbf{#1}\quad}

\newtheorem{theorem}{Theorem}

\theoremstyle{definition}
\newtheorem*{example}{Example}

\lstdefinelanguage{pddl}
{
    keywords={
        define,
    },
    morekeywords={[2]
        domain,
        problem,
        requirements,
        predicates,
        types,
        objects,
        action,
        event,
        process,
        init,
        goal,
    },
    morekeywords={[3]
        parameters,
        vars,
        precondition,
        effect,
        pr-effect,
        f\-head,
        f-exp-t,
    },
    morekeywords={[4]
        forall,
        or,
        and,
        not,
        when,
        =,
        exists,
        imply,
        increase,
        decrease,
    },
    comment=[l]{\;},
    sensitive=true
}[keywords, comments]

\usepackage{authblk}

\author[1]{Vitaliy Batusov}
\author[2]{Mikhail Soutchanski}
\affil[1]{York University, Toronto, Canada\quad \texttt{vbatusov@protonmail.com}}
\affil[2]{Ryerson University, Toronto, Canada\quad \texttt{mes@scs.ryerson.ca}}
{
    \makeatletter
    \renewcommand\AB@affilsepx{: \protect\Affilfont}
    \makeatother

}

\title{A Logical Semantics for PDDL+}

\begin{document}
 
\maketitle
\begin{abstract}
PDDL+ is an extension of PDDL2.1 which incorporates fully-featured autonomous processes and allows for better modelling of mixed discrete-continuous domains. Unlike PDDL2.1, PDDL+ lacks a logical semantics, relying instead on state-transitional semantics enriched with hybrid automata semantics for the continuous states. This complex semantics makes analysis and comparisons to other action formalisms difficult. In this paper, we propose a natural extension of Reiter's situation calculus theories inspired by hybrid automata. The kinship between PDDL+ and hybrid automata allows us to develop a direct mapping between PDDL+ and situation calculus, thereby supplying PDDL+ with a logical semantics and the situation calculus with a modern way of representing autonomous processes. We outline the potential benefits of the mapping by suggesting a new approach to effective planning in PDDL+.
\end{abstract}
\section{Introduction}

The Planning Domain Definition Language (PDDL) is the dominant AI formalism for expressing planning instances. Since its introduction by \cite{mcdermott1998}, PDDL has undergone several revisions and extensions in response to the insights 
gained from the practice of automated planning at the International Planning Competitions and elsewhere in academia and the industry. PDDL+ is a prominent offshoot of PDDL2.1 \cite{fl2003} which was introduced in \cite{fl2006} in order to significantly improve the ability to model discrete-continuous domains.

In a nutshell, PDDL+ introduces a more powerful model of continuous temporal change, namely the \emph{start-process-stop} model, which supersedes the previous \emph{durative action} model. This is motivated by an abundance of examples of domains wherein the durative actions are inappropriate or insufficient due to the complex interactions of discrete occurrences and continuous processes \cite{boddy2003,mcdermott2003}. \citeauthor{fl2006} cite the hybrid automata (HA) and the knowledge representation (KR) communities as influences for the formulation of PDDL+. The key feature borrowed by PDDL+ from the hybrid automata approach is its explicit recognition of two distinct modes of change --- the logical state transitions and the continuous change over time within a logical state. Thus, in PDDL+, processes can be initiated or terminated by any discrete change, but have continuous and possibly concurrent effects on numerical fluents. In KR, the start-process-stop model first appeared in the work of \cite{pinto-phd,ternovskaia} on situation calculus (SC), but has not been developed far enough to formulate continuous time-dependent change explicitly and achieve the accompanying representational benefits.

The formal semantics of PDDL originated with the STRIPS semantics of \cite{lifschitz1987} and reached maturity in \cite{fl2003} for PDDL2.1. As pointed out in \cite{ClassenHuLakemeyer2007}, this meta-theoretic semantics achieves its stated goals, but is difficult to grasp and is rather unwieldy for the purposes of analysis and comparison to other action formalisms. \citeauthor{ClassenHuLakemeyer2007} remedy this by providing an alternative, purely logical semantics for PDDL2.1. Their semantics is based on a subset of situation calculus called $\mathcal{ES}$, due to \cite{es}, and handles the durative action model in a robust and concise manner. In PDDL+, the semantics of continuous change is defined in terms of hybrid automata, owing to the kinship in the modelling approach and aiming for cross-pollination between the fields of automated planning and control theory. However, while this HA semantics is more amenable to analysis, it remains wrapped in the complex meta-theoretic semantics of PDDL2.1.

In this paper, we present a natural extension of SC basic action theories \cite{reiter} which aligns closely with the HA and PDDL+ approach to representing continuous change. 
Namely, in addition to the usual atemporal situation-dependent fluents, we introduce new temporal fluents such as $velocity(b,t,s)$ which explicitly depend on real-valued time $t$.
Due to the shared origins, this enables our ``hybrid'' situation calculus to provide a situation calculus semantics for PDDL+. The benefit is mutual in that PDDL+, like PDDL2.1, gets a robust logical semantics and the situation calculus gets a modern HA-like means of representing discrete-continuous change, completing the loop of cross-pollination between all three fields.  
We discuss how the ``hybrid'' SC can contribute to previous work on planning in SC to perform automated planning in PDDL+ with only minimal grounding of the action and process schemas.

\section{Hybrid Situation Calculus}

\subsection{Situation Calculus}
Situation calculus (SC) is a second-order language for representing dynamic worlds. It has basic sorts (\textsc{Situation}, \textsc{Action}, \textsc{Object}) and a rich vocabulary for constructing formulas over terms of these sorts. Reiter (\citeyear{reiter}) shows that to solve many reasoning problems about actions, it is convenient to work with \emph{basic action theories} (BATs). \cite{es} develop a variant of SC called $\mathcal{ES}$ (and an $\mathcal{ES}$-variant of BATs) which is used by \cite{ClassenHuLakemeyer2007} to supply a SC semantics for PDDL2.1. While $\mathcal{ES}$ offers syntactic and proof-theoretic benefits, it gives up the ability to express interesting statements about situations such as ``there exists a situation such that every situation preceding it satisfies $P$'' --- i.e., exactly the kind of statements about dynamic systems which control theory is interested in proving. For this reason, the present paper maps PDDL+ instances to Reiter's BATs and not to $\mathcal{ES}$.

The main ingredients of a BAT are precondition axioms and successor state axioms. For each action
function $A(\bar{x})$, an \emph{action precondition axiom} (APA) has the syntactic
form
\begin{align*}
\forall \bar{x},s.\ \ 
Poss(A(\bar{x}),s) \liff \Pi_A(\bar{x},s),
\end{align*}
meaning that the action $A(\bar{x})$ is possible in situation $s$ if and only if $\Pi_A(\bar{x},s)$ holds in $s$, where $\Pi_A(\bar{x},s)$ is a formula with free variables among $\bar{x} \eq (x_1, \ldots, x_n)$ and $s$. Situations are first order (FO) terms which denote sequences of actions. A distinguished constant $S_0$ is used to denote the \emph{initial situation}, and function $do(\alpha, \sigma)$ denotes the situation that results from performing action $\alpha$ in situation $\sigma$. Every situation corresponds uniquely to a sequence of actions. We use $do([\alpha_1,\ldots,\alpha_n],S_0)$ to denote complex situation terms obtained by consecutively performing $\alpha_1,\ldots,\alpha_n$ in $S_0$. The notation $\sigma^\prime \sqsubseteq \sigma$ means that either situation $\sigma^\prime$ is a subsequence of situation $\sigma$ or $\sigma \eq \sigma^\prime$.
The formula $\forall a \forall s^\prime(do(a,s^\prime)\sqsubseteq \sigma \to Poss(a,s^\prime))$, abbreviated as $executable(\sigma)$, captures situation $\sigma$ all of whose actions are consecutively possible. Every BAT contains a set $\Sigma$ of foundational axioms which characterize situations as a single finitely branching infinite tree starting from $S_0$. Objects are FO terms other than actions and situations that depend on the domain of application. Above, $\Pi_A(\bar{x},s)$ is a formula \emph{uniform} in situation argument $s$: it does not mention the predicates $Poss$, $\sqsubseteq$, it does not quantify over variables of sort situation, it does not mention equality on situations, and it has no occurrences of situation terms other than $s$. 
For each relational fluent $F(\bar{x},s)$ and each functional fluent $f(\bar{x},s)$, respectively, a \emph{successor state axiom} (SSA) has the form 
\begin{align*}
&F(\bar{x},do(a,s)) \liff \Phi_F(\bar{x},a,s),\\
&f(\bar{x},do(a,s))\eq y \liff \phi_f(\bar{x},y,a,s),
\end{align*}
where $\Phi_F(\bar{x},a,s)$ and $\phi_f(\bar{x},y,a,s)$ are formulas uniform in $s$, all of whose free variables are among those explicitly shown. 
(As usual, all free variables are $\forall$-quantified at front.)
In addition, a BAT $\mathcal{D}$ contains an \emph{initial theory}: a finite set $\mathcal{D}_{S_0}$ of FO formulas whose only situation term is $S_0$ (and possibly formulas without a situational argument). 
Finally, BATs
include a set $\mathcal{D}_{una}$ of unique name axioms for actions (UNA) specifying that two actions are different if their names are different and that identical actions have identical arguments. If a BAT has functional fluents, it is required to satisfy a consistency property whereby, for the right-hand side $\phi_f(\bar{x},y,a,s)$ of the SSA of each functional fluent $f$, there must exist a unique $y$ such that $\phi_f(\bar{x},y,a,s)$ is entailed by $\mathcal{D}_{S_0} \cup \mathcal{D}_{una}$.

	BATs enjoy the \emph{relative satisfiability} property \cite{pirri99}: a BAT $\mathcal{D}$ is satisfiable whenever $\mathcal{D}_{una} \cup \mathcal{D}_{S_0}$ is. This property allows one to disregard the more problematic parts of a BAT, like the second-order induction axiom, when checking satisfiability. Moreover, BATs benefit from \emph{regression}-based reasoning, a powerful and efficient reasoning mechanism for answering queries about the future (a problem known as \emph{projection}).

To accommodate time, Reiter adds a temporal argument to all action functions and introduces two special function symbols. The symbol $time: \textsc{action} \mapsto \textsc{time}$ is used to access the time of occurrence of an action via its term and is specified by an axiom $time(A(\bar{x},t)) \eq t$ (included in $\mathcal{D}_{S_0}$) for every action function $A(\bar{x},t)$ in the vocabulary of the BAT. The symbol $start : \textsc{situation} \mapsto \textsc{time}$ is used to access the starting time of situation $s$ and is specified by the new foundational axiom $start(do(a,s)) = time(a)$. The starting time of $S_0$ is not enforced, and the time points constituting the timeline with dense linear order are assumed to always have the standard interpretation (along with $+$, $<$, etc.). To outlaw temporal paradoxes, the abbreviation $executable(s)$ is redefined as
\begin{align*}
\forall a, s^\prime. do(a,s^\prime)\!\sqsubseteq\! s \to (Poss(a,s^\prime)\! \land \!start(s^\prime)\!\leq\!time(a)).
\end{align*}

To model exogenous (``coercive'') events, Reiter develops an elegant theory of \emph{natural actions} --- non-agent actions that occur spontaneously as soon as their precondition is satisfied. Such actions are marked using the predicate symbol $natural$ as a part of $\mathcal{D}_{S_0}$, e.g., $natural(bounce(Ball_1, t))$, and their semantics are encoded by a further modification of $executable(s)$ (see below). We use natural actions to induce relational change based on the values of the continuous quantities.

\subsection{Continuous Processes in Situation Calculus}

The crux of our proposal to extend BATs is to add a new kind of axioms called \emph{state evolution axioms} (SEA) to 
Reiter's successor state axioms. The successor state axioms specify, as usual,
how fluents change when actions are executed, i.e., they characterize the transitions between logical states due to actions.
The state evolution axioms specify  how  the flow of time can bring  changes 
in system parameters within a given situation while no actions are executed.
Thus, we maintain the fundamental assumption of SC that all \emph{discrete} change is due to actions, though situations now include a temporal evolution.

Our starting point is the \emph{temporal change axiom} (TCA) which describes a single law governing the evolution of a particular temporal fluent due to the passage of time in a particular context of an arbitrary situation. Each temporal fluent has time $t$ as an argument, e.g. $velocity(b,t,s)$ --- the velocity of the ball $b$ at the moment of time $t$ in situation $s$. We assume that a TCA for a temporal functional fluent $f$ has the general syntactic form
\begin{align}\label{eq:tca}
\gamma(\bar{x}, s) \land \delta(\bar{x},y,t,s) \to f(\bar{x}, t, s) \eq y,
\end{align}
where $t$, $s$, $\bar{x}$, $y$ are variables and $\gamma(\bar{x},s)$, $\delta(\bar{x},y,t,s)$ are formulas uniform in $s$ whose free variables are among those explicitly shown. We call $\gamma(\bar{x},s)$ the \emph{context} as it specifies the condition under which the formula $\delta(\bar{x},y,t,s)$ is to be used to compute the value of fluent $f$ at time $t$. Note that contexts are time-independent. 
The formula $\delta(\bar{x},y,t,s)$ encodes a function which specifies $y$ in terms of the values of other fluents at $s$, $t$. Aside from the consistency conditions below, we leave the general form of $\delta$ without constraint. 

For each TCA (\ref{eq:tca}) to be \emph{well-defined}, we require that the background theory entails
\begin{align}\label{eq:exists}
\gamma(\bar{x}, s) \to \exists y\, \delta(\bar{x},y,t,s).
\end{align}
In other words, whatever the circumstance, the TCA must supply a value for the quantity modelled by $f$ if its context is satisfied.

A set of $k$ well-defined temporal change axioms for some fluent $f$ can be equivalently expressed as an axiom 
\begin{align}\label{eq:nf}
\Phi(\bar{x},y,t,s) \to f(\bar{x}, t, s) \eq y,
\end{align}
where $\Phi(\bar{x},y,t,s)$ is $\bigvee_{1\leq i \leq k}(\gamma_i(\bar{x}, s) \land  \delta_i(\bar{x},y,t,s))$. We additionally require that the background theory entails
\begin{align}\label{eq:unique}
\Phi(\bar{x},y,t,s) \land \Phi(\bar{x},y^\prime,t,s) \to y\eq y^\prime.
\end{align}
Condition (\ref{eq:unique}) guarantees the consistency of the axiom (\ref{eq:nf}) by preventing a continuous quantity from having more than one value at any moment of time. 
With condition (\ref{eq:unique}), we can assume  w.l.o.g. that all contexts in the given set of TCA are pairwise mutually exclusive wrt the background theory $\mathcal{D}$.

Having combined all laws which govern the evolution of $f$ with time into a single axiom (\ref{eq:nf}), we can make a causal completeness assumption: \emph{there are no other conditions under which the value of $f$ can change in $s$ from its initial value at $start(s)$ as a function of $t$}. We capture this assumption formally by the explanation closure axiom
\begin{align}\label{eq:eca}
\begin{split}
&f(\bar{x}, t, s) \neq f(\bar{x},start(s), s) \to \exists y \, \Phi(\bar{x},y,t,s).
\end{split}
\end{align}

\begin{theorem}\label{th:main}
The 
conjunction of 
axioms (\ref{eq:nf}) and (\ref{eq:eca}) in the models of (\ref{eq:unique}) is logically equivalent to
\begin{align}\label{eq:sea2}
\begin{split}
&
\hspace*{-2mm}
f(\bar{x},t,s)\eq y \liff [\Phi(\bar{x},y,t,s) \lor{}\\
&\quad \quad \quad \quad \quad \quad 
	y \eq f(\bar{x},start(s), s) \land \neg \Psi(\bar{x},y,t,s)],
\end{split}
\end{align}
where $\Psi(\bar{x},s)$ denotes $\bigvee_{1\leq i \leq k}\gamma_i(\bar{x}, s)$.
\end{theorem}

We call the formula (\ref{eq:sea2}) a \emph{state evolution axiom} (SEA) for the fluent $f$. 
Note what the SEA says: $f$ evolves with time during $s$ according to some law whose context is realized in $s$ or stays constant if no context holds in $s$.  
Our causal completeness assumption (\ref{eq:eca}) applies both to physical quantities 
and  to their derivatives. This assumption simply states that all reasons for
change have been already accounted for in  (\ref{eq:nf}).
The proof of this theorem is available in \cite{arxiv:abs-1807-04861}.
It is important to realize that $\mathcal{D}_{se}$, a set of SEAs, 
complements the SSAs that are derived in \cite{reiter1991} using a similar technique.

\subsection{``Hybrid'' Basic Action Theories}

The SEA for some temporal fluent $f$ does not completely specify the behaviour of $f$ because it talks only about change within a single situation $s$. To complete the picture, we need a SSA describing how the value of $f$ changes (or does not change) when an action is performed. A straightforward way to 
accomplish this would be by an axiom which would enforce continuity of value.  However, this choice would preclude the ability to model action-induced discontinuous jumps in the value of the continuously varying quantities or their derivatives, such as the sudden change of acceleration from $0$ to $-9.8 m/s^2$ when an object is dropped. To circumvent this limitation, for each temporal functional fluent $f(\bar{x}, t, s)$, we introduce an auxiliary atemporal functional fluent $f\init(\bar{x},s)$ whose value in $s$ represents the value of the physical quantity modelled by $f$ in $s$ at the time instant $start(s)$. We axiomatize $f\init$ using a SSA derived from the axioms
\begin{align*}
&e(\bar{x},y,a,s) \to f\init(\bar{x}, do(a,s)) = y,\\\
&\neg\exists y(e(\bar{x},y,a,s)) \to f\init(\bar{x}, do(a,s)) \eq f(\bar{x},time(a),s),
\end{align*}
where the former is a Reiter's effect axiom in normal form and the latter asserts that if no relevant effect is invoked by the action $a$, $f\init$ assumes the most recent value of the continuously evolving fluent $f$. This latter axiom enforces temporal continuity in the value of $f$ in the case when there is no reason for change.

 The SSA for $f\init$ has the general syntactic form
\begin{align}\label{eq:init}
\begin{split}
&f\init(\bar{x}, do(a,s)) \eq y \liff \Omega(\bar{x}, y, a, s),
\end{split}
\end{align}
where $\Omega(\bar{x},y, a, s)$ is a formula uniform in $s$ 
whose purpose is to describe how the initial value of $f$ in $do(a,s)$ relates to its value at the same time instant in $s$ (i.e., prior to $a$). To establish a consistent relationship between temporal fluents and their atemporal \emph{init}-counterparts, we require that, in an arbitrary situation, the continuous evolution of each temporal fluent $f$ starts with the value computed for $f\init$ by its successor state axiom.

A \emph{hybrid basic action theory} is a collection of axioms $\mathcal{D} = \Sigma \cup \mathcal{D}_{ss} \cup \mathcal{D}_{ap} \cup \mathcal{D}_{una} \cup \mathcal{D}_{S_0} \cup \mathcal{D}_{se}$ such that
\begin{enumerate}\setlength\itemsep{0.1em}
\item Every action symbol mentioned in $\mathcal{D}$ is temporal;
\item $\Sigma \cup \mathcal{D}_{ss} \cup \mathcal{D}_{ap} \cup \mathcal{D}_{una} \cup \mathcal{D}_{S_0}$ constitutes a BAT as per  Definition 4.4.5 in \cite{reiter};

\item $\mathcal{D}_{se}$ is a set of state evolution axioms of the form
\begin{align}\label{eq:seagf}
f(\bar{x},t,s) \eq y \liff \psi_f(\bar{x},t,y,s)
\end{align}
where $\psi_f(\bar{x},t,y,s)$ is uniform in $s$, such that for every temporal functional fluent $f$ in $\mathcal{D}_{se}$, $\mathcal{D}_{ss}$ contains an additional SSA of the form (\ref{eq:init}) for $f\init$;
\item For each SEA of the form (\ref{eq:seagf}), the following consistency properties must be entailed by $\mathcal{D}_{una} \cup \mathcal{D}_{S_0}$:
\begin{align}\label{eq:func-consistency}
\begin{split}
&\quad \forall \bar{x} \forall t. \;\exists y (\psi_f(\bar{x}, t, y, s)) \land{} \\
&\quad \forall y \forall y^\prime (\psi_f(\bar{x}, t, y, s) \land \psi_f(\bar{x}, t, y^\prime, s) \to y \eq y^\prime),
\end{split}
\end{align}
\begin{align}\label{eq:init-consistency}
&\quad \exists y ( f\init(\bar{x},s) \eq y \land{} \psi_f(\bar{x},start(s), y, s)); 
\end{align}
\end{enumerate}

It can be shown that ``hybrid'' BATs retain a variant of the relative satisfiability property, and a variant of the regression operator can also be defined.

\subsection{Differential Equations in Situation Calculus}

Both science and engineering describe dynamic systems in terms of differential equations. Hybrid automata are a prime example of a formalism which relies on ordinary differential equations (ODE)  to specify continuous behaviour. 
PDDL+ specifies continuous effects using an explicit ODE of the form $\der{f(t)}{t} = G(t,f)$, where $G(t,f)$ is a continuously differentiable, or more generally, a Lipschitz continuous function over functional fluents and constants. An initial value $f(t_0)$ is implicitly given, and together with the ODE it defines the \textit{initial value problem} that thanks to assumptions about $G$ has a unique solution, e.g., see \cite{teschl2012ordinary}. 
We encode PDDL+  process effects using the equivalent integral form $f(t) = \int_{t_0}^t G(\tau,f)\; d\tau + f(t_0)$.

Henceforth, we require that all FOL structures interpret the definite integral symbol in the standard way. Let $\mathcal{M}$ be an arbitrary situation calculus structure, $\mu$ be an object assignment, $h(\bar{x},t,s)$ a SC term of sort $\mathbb{R}$ whose free variables are among $\bar{x},t,s$ (of sorts \textsc{Object}, $\mathbb{R}$, \textsc{Situation}, resp.), and let $\tau_1, \tau_2$ be terms of sort $\mathbb{R}$. Then we require that
\begin{align*}
\int_{\tau_1}^{\tau_2}\!\! h(\bar{x}, t, s) \,dt\, \strut^\mathcal{M}[\mu] = \int_{\tau_1^\mathcal{M}[\mu]}^{\tau_2^\mathcal{M}[\mu]} \!h^\mathcal{M}(\bar{x}^\mathcal{M}[\mu], t, s^\mathcal{M}[\mu]) \,dt.
\end{align*}
The modeller must ensure that $h^\mathcal{M}(\bar{x}^\mathcal{M}[\mu], t, s^\mathcal{M}[\mu])$ is a continuous real-valued function defined on the interval $[\tau_1^\mathcal{M}[\mu], \tau_2^\mathcal{M}[\mu]]$.

\section{A mapping from PDDL+ to SC}

Our mapping explicitly captures all PDDL+ features subsumed by the requirements \ttt{:adl} + \ttt{:fluents}. 

A \emph{PDDL+ planning instance} is a pair $I = (Dom, Prob)$ where $Dom = (\mathcal{T}$, $\mathcal{C}$, $\mathcal{F}$, $\mathcal{R}$, $\mathcal{A}$, $\mathcal{E}$, $\mathcal{P})$ describes the domain dynamics and $Prob = (\mathcal{O}, \mathcal{I}, \mathcal{G})$ describes problem-specific objects, the initial state, and the goal state. A planning instance is formulated according to a formal grammar found in \cite{fl2003,fl2006}. In short, $\mathcal{F}$ is a set of \emph{function signatures}, $\mathcal{R}$ is a set of \emph{predicate signatures}, $\mathcal{O}$ and $\mathcal{C}$ are sets of \emph{objects} which serve as arguments to functions and predicates, $\mathcal{T}$ is a \emph{type hierarchy} over a set of \emph{type names}, $\mathcal{A}$ is a set of \emph{instantaneous action schemas}, $\mathcal{E}$ is a set of \emph{event schemas}, $\mathcal{P}$ is a set of \emph{process schemas}, $\mathcal{I}$ is the \emph{initial state specification}, and $\mathcal{G}$ is a \emph{goal state specification}. Below, we go over these sets in detail.

\lab{Names}A PDDL+ instance $I$ uses the set of alphanumeric strings as a namespace for types, objects, predicates, actions, etc. In the grammar of PDDL+, the syntactic category \texttt{<name>} refers to any member of this namespace. Our translation uses the same namespace for the vocabulary of BAT. 

\lab{Objects and Constants}The instance $I$ supplies a finite set $\mathcal{O}$ of \emph{objects} --- symbols for distinct primitive entities operated upon --- and a finite set of \emph{constants}, which are objects hard-coded into the domain and shared between all planning problems for that domain. In SC, the elements of $\mathcal{O} \cup \mathcal{C}$ correspond to constants with unique name axioms and a domain closure axiom. 

\lab{Types}PDDL+ has two basic types, \texttt{object} and \texttt{number}. Type \texttt{object} acts as a parent type for domain-specific types of objects. The type \texttt{number} is not extensible and is only used as the return type of functions. A finite set of object types is specified as a part of the type hierarchy $\mathcal{T}$.

In our mapping, the types \ttt{object} and \texttt{number} are not modelled explicitly and instead correspond to terms of the sorts \textsc{Object} and $\mathbb{R}$, respectively. Each defined type corresponds to a unary predicate of the same name whose argument is of sort \textsc{Object}. 
In slight abuse of notation, we write $T \in \mathcal{T}$ to denote a primitive type name occurring in $\mathcal{T}$.

The type hierarchy $\mathcal{T}$ is specified by an instance of the syntactic category $\ttt{<typed list (name)>}$, i.e., an expression of the form $\ttt{(}T_1\ttt{ - }TE_1 \ldots T_n\ttt{ - }TE_n \ttt{)}$ where each $T_i$ is a primitive type name and each $TE_i$ is either a primitive type name or a structure of the form $\ttt{(either } Y_1 \ldots Y_k\ttt{)}$ where $k \geq 1$ and $Y_j$ are primitive type names. We translate $\mathcal{T}$ to SC as follows. For each $1 \leq i \leq n$, we include an axiom $\forall x (\widehat{T_i} \to \widehat{TE_i})$, with the mapping function $\widehat{\cdot}$ defined in Table \ref{t:hat}. As each object $O \in \mathcal{O} \cup \mathcal{C}$ is supplied with its own type expression $TE$, we include an axiom $\widehat{TE}({O})$ to associate $O$ with its type.

\lab{Variables}PDDL+ variables are names prepended with a question mark, used as placeholders for the objects in action/event/process schemas and goal description quantifiers. We map variables to SC variables according to Table \ref{t:hat}.

\lab{Predicates}$\mathcal{R}$ is a non-empty finite set of predicate signatures of the form $\ttt{(} R \ttt{ ?}x_1 \ttt{ - }TE_1 \ldots \ttt{?}x_n \ttt{ - }TE_n\ttt{)}$ where $R$ is the predicate name, $\ttt{?}x_i \ldots \ttt{?}x_n$ are variable names, and $TE_i$ is a type expression for the $i$-th argument of $R$. As before, the type expression $TE_i$ is either the name of a primitive type from $\mathcal{T}$ or a structure of the form $\ttt{(either } T_1 \ldots T_k\ttt{)}$ where $k \geq 1$ and $T_i \in \mathcal{T}$ for $1 \leq i \leq k$. Let $arity(R)$ denote $n$, the arity of $R$ as specified by $\mathcal{R}$.

In PDDL+, the dynamic nature of predicates is dictated by the effects of actions/events/processes. In SC, we make this distinction formal. We partition $\mathcal{R}$ into the mutually exclusive sets $\mathcal{R}_s$ (``static'') and $\mathcal{R}_d$ (``dynamic'') such that $\mathcal{R}_d = \{R \in \mathcal{R} \mid R \text{ appears in effects of some action/event}\}$ and $\mathcal{R}_s = \mathcal{R} \setminus \mathcal{R}_d$. The predicate names in $\mathcal{R}_s$ become SC static (rigid) predicates with the signature $\textsc{Object}^{arity(R)}$. The symbols in $\mathcal{R}_d$ become SC relational fluents with the signature $\textsc{Object}^{arity(R)} \times \textsc{Situation}$. 

\lab{Functions}$\mathcal{F}$ is  a possibly empty finite set of function signatures of the form $\ttt{(} F \ttt{ ?}x_1 \ttt{ - }TE_1 \ldots \ttt{?}x_n \ttt{ - }TE_n\ttt{)}$ where $F$ is the function name, $\ttt{?}x_1 \ldots \ttt{?}x_n$ are variable names, and $TE_i$ is the type description of the $i$-th argument of $F$ as in the case for predicates. Let $arity(F)$ denote the arity of $F$ as specified by $\mathcal{F}$. As with predicates, we partition PDDL+ functions into classes based on how they change. 
$\mathcal{F}_t$ (``temporal'') are exactly those members $F$ of $\mathcal{F}$ such that $\ttt{(increase (}  F \ttt{ \ldots) \ldots)}$ occurs in some process schema $\mathcal{P}$, i.e., the functions which undergo continuous evolution. The function signatures in $\mathcal{F}_t$ correspond to SC temporal fluents with the signature $\textsc{Object}^{arity(R)} \times \mathbb{R} \times \textsc{Situation} \mapsto \mathbb{R}$. Additionally, each member of $\mathcal{F}_t$ gives rise to a related fluent $F\init$ with the signature $\textsc{Object}^{arity(R)} \times \textsc{Situation} \mapsto \mathbb{R}$. $\mathcal{F}_d$ (``dynamic'') are those members $F \in (\mathcal{F} \setminus \mathcal{F}_t)$ which can be affected by some instantaneous event or action from $\mathcal{A} \cup \mathcal{E}$. The members of $\mathcal{F}_d$ correspond to SC functional fluents with the signature $\textsc{Object}^{arity(F)} \times \textsc{Situation} \mapsto \mathbb{R}$. Finally, $\mathcal{F}_s = \mathcal{F} \setminus (\mathcal{F}_d \cup \mathcal{F}_t)$ are the inherently static functions which are mapped to SC static functions with the signature $\textsc{Object}^{arity(F)} \mapsto \mathbb{R}$.  We write $\widehat{\mathcal{F}}$, $\widehat{\mathcal{F}}_t$, etc. to denote sets of SC terms which correspond to the respective PDDL+ function signatures.

\lab{Goal Descriptions}The category \texttt{<GD>} (\emph{goal description}) represents a FOL-like statement used to describe an action precondition or a goal condition. The grammar of \texttt{<GD>} defines it as an expression constructed from predicate and (in)equality atoms on functions (but not process atoms) using standard logical connectives and quantifiers. Table \ref{t:hat} summarizes the mapping of arbitrary goal descriptions $X$ to the expressions $\widehat{X}$ of situation calculus. It can be easily shown that \ 
(a) $\exists t(t \geq start(s) \land \widehat{X})$ is a well-formed SC formula uniform in $s$, and \ 
(b) whenever a PDDL+ state and a SC structure with an object assignment agree on the values of all terms and atoms, they interpret $X$ and $\exists t(t \geq start(s) \land \widehat{X})$ (respectively) identically.

\begin{table}[h]
\caption{A mapping from PDDL+ to SC expressions. On the right, $x$ is a variable of sort \textsc{Object}, $y$ a variable of sort $\mathbb{R}$, $t$ a variable of sort $\mathbb{R}$ (for time), and $s$ is a variable of sort \textsc{Situation}. Here and subsequently, $\phi(x/x_i)$ denotes the result of replacing $x$ by $x_i$ in $\phi$.}\label{t:hat}
\small
\begin{tabular}{ l | l}
\toprule
$X$ & $\widehat{X}$ \\
\midrule
$\ttt{?}x$ & variable $x$ of sort \textsc{Object}\\
$O \in \mathcal{O} \cup \mathcal{C}$ & constant $O$ of sort \textsc{Object}\\
 numeric literal $n$ & $n$\\
 $T \in \mathcal{T}$ & $T(x)$ \\
 $\ttt{(either} \; T_1 \ldots T_n\ttt{)}$ & $(T_1(x) \lor \ldots \lor T_n(x))$ \\
$\ttt{(}R\; b_1 \ldots b_n\ttt{)}, R \in \mathcal{R}_s$ & $R(\widehat{b}_1 \cdots \widehat{b}_n)$ \\
 $\ttt{(}R\; b_1 \ldots b_n\ttt{)}, R \in \mathcal{R}_d$ & $R(\widehat{b}_1 \cdots \widehat{b}_n, s)$ \\
 $\ttt{(}F\; b_1 \ldots b_n\ttt{)}, F \in \mathcal{F}_s$ & $F(\widehat{b}_1 \cdots \widehat{b}_n)$ \\
 $\ttt{(}F\; b_1 \ldots b_n\ttt{)}, F \in \mathcal{F}_d$ & $F(\widehat{b}_1 \cdots \widehat{b}_n, s)$ \\
 $\ttt{(}F\; b_1 \ldots b_n\ttt{)}, F \in \mathcal{F}_t$ & $F(\widehat{b}_1 \cdots \widehat{b}_n, t, s)$ \\
 $\ttt{(and} \; Y_1 \ldots Y_n\ttt{)}$ &  $(\widehat{Y}_1 \land \ldots \land \widehat{Y}_n)$ \\
 $\ttt{(or} \; Y_1 \ldots Y_n\ttt{)}$ &  $(\widehat{Y}_1 \lor \ldots \lor \widehat{Y}_n)$ \\
 $\ttt{(not} \; Y\ttt{)}$ &  $\neg (\widehat{Y})$ \\
 $\ttt{(imply} \; Y_1 \; Y_n\ttt{)}$ &  $(\widehat{Y}_1) \to (\widehat{Y}_n)$ \\
 \makecell[tl]{$\ttt{(exists (?}x_1 \ttt{-} TE_1 $\\\qquad$\ldots \ttt{?}x_n \ttt{-} TE_n\ttt{) } Y \ttt{)}$} & \makecell[tl]{$\exists x_1 \ldots \exists x_n$\\ $(\bigwedge_{1\leq i \leq n}\widehat{TE}_i(x/x_i) \land (\widehat{Y}))$}\\
 \makecell[tl]{$\ttt{(forall (?}x_1 \ttt{-} TE_1 $\\ \qquad $\ldots \ttt{?}x_n \ttt{-} TE_n\ttt{) } Y \ttt{)}$} &  \makecell[tl]{$\forall x_1 \ldots \forall x_n$\\ $(\bigwedge_{1\leq i  \leq n} \widehat{TE}_i(x/x_i) \to (\widehat{Y}))$} \\
 $\ttt{(- } Y \ttt{)}$ & $- (\widehat{Y})$ \\
 \makecell[tl]{$\ttt{(}\circ Y_1 \; Y_2\ttt{)}, \circ \in \{+,-,\times,$\\ \quad$\div,>,<,=,\leq, \geq\}$} & $(\widehat{Y}_1 \circ \widehat{Y}_n)$ \\ 
\bottomrule
\end{tabular}
\end{table}

\begin{table}[h]
\caption{An auxiliary mapping from instances of \ttt{<effect>} for a PDDL+ action $A(\bar{z})$ to SC expressions. $A(\bar{z},t)$ is a SC action term corresponding to the PDDL+ action $A(\bar{z})$. Let $RelOp$ be $\{\ttt{increase}, \ttt{decrease},  \ttt{scale-up}, \ttt{scale-down}\}$.}\label{t:effect}
\small
\begin{tabular}{l|l}
\toprule
$X$ & $\widetilde{X}$ \\
\midrule
$\ttt{(}R\; b_1 \ldots b_n\ttt{)}, R \in \mathcal{R}_d$ & $\exists t R(\widehat{b}_1 \cdots \widehat{b}_n, do(A(\bar{z},t),s))$ \\
$\ttt{(}F\; b_1 \ldots b_n\ttt{)}, F \in \mathcal{F}_d$ & $\exists t F(\widehat{b}_1 \cdots \widehat{b}_n, do(A(\bar{z},t),s))$ \\
$\ttt{(}F\; b_1 \ldots b_n\ttt{)}, F \in \mathcal{F}_t$ & $\exists t F\init(\widehat{b}_1 \cdots \widehat{b}_n, do(A(\bar{z},t),s))$ \\
 \makecell[tl]{$\ttt{(forall}$\\\qquad $\ttt{(?}x_1 \ldots \ttt{?}x_n\ttt{)}\; Y \ttt{)}$} &  $\forall x_1 \cdots \forall x_n (\widetilde{Y})$ \\
$\ttt{(when }Y_1 \; Y_2\ttt{)}$ &  $ (\widehat{Y}_1 \to \widetilde{Y}_2)$ \\
$\ttt{(assign } Y_1 \; Y_2\ttt{)}$ & $\forall y(\widehat{Y}_2 \eq y \to \widetilde{Y}_1 \eq y)$ \\
$\ttt{(}\circ Y_1 \; Y_2\ttt{)}, \circ \in RelOp$ & $\forall y(\widehat{Y}_1 \,\widetilde{\circ}\, \widehat{Y}_2 \eq y \to \widetilde{Y}_1 \eq y)$ \\
$\circ \in RelOp$ & $+/-/\times/\div$ \\
$X$ not appearing above & $\widehat{X}$\\
\bottomrule
\end{tabular}
\end{table}

\lab{Actions}$\mathcal{A}$ is a set of schemas of the form
\begin{lstlisting}
(:action $A$
:parameters  (?$x_1$ - $TE_1$ ... ?$x_n$ - $TE_n$)
:precondition $Pre_A$
:effect $\Eff_A$)
\end{lstlisting}
where $A$ is the action name, $Pre_A$ is an instance of \ttt{<GD>}, and $\Eff_A$ is an instance of \ttt{<effect>}. According to \cite{mcdermott1998}, all variables that occur in a schema must either appear under \ttt{:parameters} or be quantified.
Each such schema generates a SC action $A(x_1, \ldots, x_n,t)$ with the mandatory axiom $time(A(\bar{x},t)) = t$ and a unique name axiom $A(\bar{u}) \eq A(\bar{v}) \to \bar{u}\eq \bar{v}$. The precondition $Pre_A$ and the parameter types $TE_i$, when translated according to Table \ref{t:hat}, yield a SC precondition axiom
\begin{align*}
Poss(A(x_1, \ldots, x_n,t), s) \liff \textstyle\bigwedge_{1 \leq i \leq n} \widehat{TE}_i(x/x_i) \land \widehat{Pre}_A.
\end{align*}
It is clear that the right-hand side of this axiom is uniform in $s$ and contains no free variables except for those shown.

Action effects are specified by the category \texttt{<effect>}, but are usually treated as sets of ground atoms, produced by explicit flattening and grounding as per Definition 5 in \cite{fl2003}. Fortunately, we can  avoid grounding in favour of conciseness. Let $\widetilde{\cdot}$ be a mapping from PDDL+ to SC expressions as defined in Table \ref{t:effect}.

\begin{theorem}
Given an instance $\Eff_A$ of \ttt{<effect>}, $\widetilde{\Eff}_A$ is a SC formula which is logically equivalent to a conjunction of axioms of the following forms: 
\begin{align*}
&  a\eq A(\bar{x},t) \land \Psi(\bar{x}, \bar{z},t,s) \to  P(\bar{x},\bar{z},do(a,s)), \\
&a\eq A(\bar{x},t) \land \Psi(\bar{x}, \bar{z},t,s) \to  \neg P(\bar{x},\bar{z},do(a,s)),\\
&a\eq A(\bar{x},t) \land \Psi(\bar{x}, y, \bar{z},t,s) \to  F(\bar{x},\bar{z},do(a,s)) \eq y.
\end{align*}
\end{theorem}
Such axioms are proper SC effect axioms that can be compiled into SSA in the usual way following Reiter's solution to the frame problem \cite{reiter1991}. Note that whenever no action or event (see below) affects a function $F$ from $\mathcal{F}_t$, the above method fails to supply the fluent $F\init$ with a SSA. Thus, for each $ F(\bar{x},t,s) \in \widehat{\mathcal{F}_t}$ unaffected by actions or events, our BAT must include a trivial SSA $F\init(\bar{x},do(a,s)) \liff  F(\bar{x},time(a),s)$.

It can be shown that the PDDL+ notion of an \emph{applicable} action coincides with the SC notion of an \emph{executable} action. Furthermore, whenever a PDDL+ state and a SC structure with an object assignment agree on the values of all terms and atoms, they also agree on the interpretation of the successor state, as determined by $\mathcal{A}$ and the SSA obtained from the axioms above, respectively.

\lab{Events} PDDL+ events are described by a set of schemas whose grammar coincides with that of action schemas and are therefore mapped to SC in the same way. However, events have a different semantics in that they occur spontaneously. To capture this semantics, we appeal to Reiter's notion of natural actions and declare each SC action $E(\bar{x},t)$ which originates from a PDDL+ event as a natural one using the axiom $natural(E(\bar{x},t))$. Since Reiter defines natural actions only for theories with true concurrency which PDDL+ lacks, we redefine $executable(s)$ accordingly:
\begin{align*}
&\forall a, s'\big[ do(a,s')\!\sqsubseteq\! s \to Poss(a, s') \!\land\! start(s') \lleq time(a) \big] \land\\
&\forall a,a', s'\big[ natural(a) \land Poss(a, s') \land do(a',s') \sqsubseteq s \to{}\\
&\hspace{4.5em} a \eq a' \lor (natural(a') \land time(a') \lleq time(a)) \big].
\end{align*}
Here, natural actions occur as soon as possible (at the true time instant $t$), but in some arbitrary order such that no agent action can appear in between two natural actions which occur at the same time instant. 

\lab{Processes}$\mathcal{P}$ is a set of schemas of the form
\begin{lstlisting}
(:process $P$
:parameters  (?$x_1$ - $TE_1$ ... ?$x_n$ - $TE_n$)
:precondition $Pre_P$
:effect $\Eff_P$)
\end{lstlisting}
where $P$ is the process name, $Pre_P$ is an instance of \ttt{<GD>}, and $\Eff_P$ is an instance of \ttt{<process-effect>}. Each such schema generates the following SC symbols:
\begin{itemize}
\item a relational fluent $P(x_1, \ldots, x_n,s)$ (denoted $\widehat{P}$) and
\item actions $begin_{P}(x_1, \ldots, x_n,t)$ and $end_{P}(x_1, \ldots, x_n,t)$.
\end{itemize}

The process-induced relational fluents $P(x_1, \ldots, x_n,s)$ are to compose the SEA contexts which determine how continuous quantities evolve based on which processes are active in the given state. The actions are natural and act as events which initiate and terminate the process in an explicit syntactic representation of the \emph{start-process-stop} model. The following axioms model this behaviour.
\begin{align*}
& Poss(begin_P(\bar{x},t),s) \liff \textstyle\bigwedge_{1 \leq i \leq n} \widehat{TE}_i(x/x_i) \land \widehat{Pre}_P,\\
& Poss(end_P(\bar{x},t),s) \liff \textstyle\bigwedge_{1 \leq i \leq n} \widehat{TE}_i(x/x_i) \land \neg \widehat{Pre}_P,\\
& P(\bar{x},do(a,s)) \liff{}\\
&\quad \exists t(a \eq begin_P(\bar{x},t)) \lor P(\bar{x}, s) \land \neg \exists t(a \eq end_P(\bar{x},t)).
\end{align*}

Process effects are specified by an instance $\Eff_P$ of the category \texttt{<process-effect>}. Without loss of generality, we consider the following simplified grammar:
\begin{lstlisting}
<process-effect> ::= (and <p-eff-t>$^\ttt{*}$)
<p-eff-t> ::= (increase <f-head> <f-exp-t>)
<f-head> ::= (<function-symbol> <term>$^\ttt{*}$)
<f-exp-t> ::= (* #t <f-exp>)
\end{lstlisting}

Crucially, each process effect defined thereby is \emph{relative}, in that it describes but a part of the continuous evolution of a quantity, the part induced by $P$ alone. To achieve axiomatic causal completeness, our mapping must combine effects due to \emph{all} processes into a universal temporal change axiom and, subsequently, SEA for each continuous fluent.

Recall that $\mathcal{F}_t$ are exactly the symbols that instantiate \ttt{<f-head>} in \ttt{<p-eff-t>} above. 
Consider the set
\begin{align*}
&Terms_P = \{\widehat{X} \mid \ttt{(increase }  X \ttt{ (...))} \in \Eff_P\}.
\end{align*}
of all function terms affected by process $P$.
For each $\widehat{X} \in Terms_{P}$ where the function name is $F$, let $\theta_{\widehat{X},P}$ be an arbitrary one-to-one variable substitution from the object variables of $\widehat{P}$ to the set $\{x_1, \ldots, x_{arity(P)}\}$ such that $\widehat{X}\theta_{\widehat{X},P} = F(x_1, \ldots, x_{arity(F)}, t, s)$. For each $\widehat{X} \in Terms_P$, consider the set $Parts_{\widehat{X},P}$ of tuples $(\widehat{P}\theta_{\widehat{X},P}, \widehat{Y}\theta_{\widehat{X},P})$ such that $\ttt{(increase }  X \ttt{ (* \#t }Y\ttt{))}$ occurs in $\Eff_P$. Each pair in this set describes a context and a continuous effect that it induces on $\widehat{X}$, normalized wrt the arguments of the term $\widehat{X}$.
For each such tuple $T = (P(\bar{z},s), \phi(\bar{z}',t,s))  \in Parts_{\widehat{X},P}$, according to the grammar, we have $\bar{z}' \subseteq \bar{z}$  and $\bar{x} \subseteq \bar{z}$. Let $\bar{w}$ denote the variables contained in $\bar{z}$ but not in $\bar{x}$. Let $Gnd(T)$ denote the set of all tuples which can be obtained from the tuple $T$ by simultaneously replacing all variables $\bar{w}$ with constants from $\mathcal{O} \cup \mathcal{C}$ in all possible ways (thereby eliminating $\bar{w}$). Let $GndParts_{\widehat{X},P} = \{Gnd(T) \mid T \in Parts_{\widehat{X},P}\}$.
Finally, for each $F \in \mathcal{F}_t$, we combine all such tuples due to all processes which affect $F$ in a single set:
\begin{align*}
B_F = \textstyle\bigcup_{P \in \mathcal{P}, \widehat{X} \in Terms_F} GndParts_{\widehat{X},P}.
\end{align*}

We can now generate the complete TCA in normal form for the fluent $F$, where $2^{B_F}$ is the power set of $B_F$: 
\begin{align*}
&\bigvee_{b_F \in 2^{B_F}}\Big[ \bigwedge_{\langle X,Y\rangle \in b_F} (X) \land \bigwedge_{\langle X,Y\rangle \in B_F \setminus b_F} (\neg X) \land{}\\
&\quad y \eq F_{init}(\bar{x},s) + \int_{start(s)}^{t} \sum_{\langle X,Y\rangle \in b_F} Y(t/t') \; dt' \Big]\\
&\qquad \to F(\bar{x},t,s) \eq y.
\end{align*}
By construction, exactly one context of this TCA is satisfied in any given FOL structure. Thus, the TCA is causally complete and can be transformed into a correct SEA by simply making the implication bidirectional.

\lab{Durative Actions}As shown in Appendix D of \cite{fl2006}, durative actions can be seen in PDDL+ as syntactic sugar for a combination of two agent actions, an event, and a process, all of which we capture above. Thus, indirectly, the mapping supports the requirement \ttt{:durative-actions}. 

\lab{Timed Initial Intervals}Timed initial intervals (TIL) first appeared in PDDL2.2 \cite{pddl2.2} and are not a part of the original PDDL+ proposal, but commonly occur in practice. They offer a means to concisely express a special kind of exogenous events which are destined to occur at prespecified times. We capture these using special-purpose natural actions---one per time point---and add the obvious effect axioms to the SSA of the fluents involved. Thus, our mapping also supports \ttt{:timed-initial-intervals}.

\lab{Initial State}The PDDL+ initial state is specified by a set of ground atoms and equalities between ground terms and numbers. These are trivially translated to SC using Table \ref{t:hat} while substituting $S_0$ for the situation term, 0 for the time argument, and asserting that $start(S_0) \eq 0$.

\lab{Goal State}A PDDL+ goal state is an arbitrary instance of \ttt{<GD>} already handled by the mapping, yielding a formula $Goal(s,t)$. To ensure that all timed initial intervals get a chance to occur, we use the time $T$ of the latest TIL (or $T \eq start(S_0)$ if no TILs are given) to constrain the goal formula: $\exists s \exists t(t \geq start(s) \land t \geq T \land Goal(s,t))$.

This completes the translation and yields a hybrid BAT $\mathcal{D}$ which concisely characterizes both a planning instance and its Tarskian semantics. Consequently, it can be directly tested for consistency and satisfiability via automated theorem proving. The PDDL+ notion of a plan corresponds to a SC ground situation, i.e., a sequence of ground actions, which are already annotated with timestamps by virtue of having a temporal argument. The planning task is then defined as finding a ground situation term $\sigma$ over the vocabulary of $\mathcal{D}$ such that $\mathcal{D} \models executable(\sigma) \land \exists t(t \geq start(\sigma) \land t \geq T \land Goal(\sigma,t))$. Here, $executable(\sigma)$ ensures that all preconditions, including those of natural actions, are met.

\begin{example}
For illustration, consider a representative fragment of the Car domain from \cite{piotrowski2016}, which models the movement of a car. The domain symbols are mapped as follows (we shorten the names for brevity).  
\begin{align*}
\texttt{running} \mapsto{} & run(s) & \in \mathcal{F}_d\\
\texttt{goal\_reached} \mapsto{} & goalr(s) & \in \mathcal{F}_d\\
\texttt{engineBlown} \mapsto{} & engb(s) & \in \mathcal{F}_d\\
\texttt{accelerate} \mapsto{} & accel(t) & \in \mathcal{A}\\
\texttt{decelerate} \mapsto{} & decel(t) & \in \mathcal{A}\\
\texttt{stop} \mapsto{} & stop(t) & \in \mathcal{A}\\
\texttt{engineExplode} \mapsto{} & expl(t) & \in \mathcal{E}\\
\texttt{a} \mapsto{} & A(s) & \in \mathcal{F}_d\\
\texttt{up\_limit} \mapsto{} & up\_limit & \in \mathcal{F}_s\\
\texttt{down\_limit} \mapsto{} & down\_limit & \in \mathcal{F}_s\\
\texttt{moving} \mapsto{} & m(s) & \in \mathcal{P}\\
\texttt{windResistance} \mapsto{} & w(s) & \in \mathcal{P}\\
\texttt{v} \mapsto{} & V(t,s) & \in \mathcal{F}_t\\
& V\init(s) & \text{auxiliary}\\
\texttt{d} \mapsto{} & D(t,s) & \in \mathcal{F}_t\\
& D\init(s) & \text{auxiliary}
\end{align*}
Additionally, the mapping extends the vocabulary with the natural actions $begin_m(t)$, $end_m(t)$, $begin_w(t)$, $end_w(t)$ which initiate and terminate the processes. The action preconditions are as follows:
\begin{align*}
&Poss(accel(t),s) \liff run(s) \land A(s) < up\_limit,\\
&Poss(decel(t),s) \liff run(s) \land A(s) > down\_limit,\\
&Poss(stop(t),s) \liff V(t,s) \eq 0 \land D(t,s) \geq 30,\\
&Poss(expl(t),s) \liff run(s) \land A(s) \geq 1 \land V(t,s) \geq 100,
\end{align*}
\begin{align*}
&Poss(begin_m(t),s) \liff run(s),\\
&Poss(end_m(t),s) \liff \neg run(s),\\
&Poss(begin_w(t),s) \liff run(s) \land V(t,s) \geq 50,\\
&Poss(end_w(t),s) \liff \neg (run(s) \land V(t,s) \geq 50).
\end{align*}
The latter four actions are natural and are asserted as such, e.g., by the axiom $natural(end_w(t))$.

Relational atemporal predicates get the SSA compiled from action effects and implicit frame axioms:
\begin{align*}
run(do(a,s)) \liff{}& run(s) \land \neg \exists t[a \eq expl(t)],\\
goalr(do(a,s)) \liff{}& \exists t[a \eq stop(t)] \lor goalr(s),\\
engb(do(a,s)) \liff{}& \exists t[a \eq expl(t)] \lor engb(s).
\end{align*}
Process fluents are also atemporal and so are described similarly, e.g. $m(do(a,s)) \liff \exists t[a \eq begin_m(t)] \lor m(s) \land \neg \exists t[a \eq end_m(t)]$.

The action effects for $\ttt{a} \mapsto \!A(s)$ yield SC effect axioms
\begin{align*}
\exists t [a \eq accel(t) \land y \eq A(s) + 1] \to A(do(a,s)) \eq y,\\
\exists t [a \eq decel(t) \land y \eq A(s) - 1] \to A(do(a,s))\eq y,\\
\exists t [a \eq explode(t) \land y \eq 0] \to A(do(a,s))\eq y,
\end{align*}
which are likewise compiled into a SSA 
\begin{align*}
A(do(a,s))\eq y \liff \phi(y,a,s) \lor y \eq A(s) \land \neg \exists y' \,\phi(y',a,s),
\end{align*}
where $\phi(y,a,s)$ is the disjunction of the left-hand sides of the effect axioms.

Although acceleration, velocity, and distance are all continuous quantities, only the latter two undergo continuous evolution in the Car domain. Velocity changes trivially across situations:
\begin{align*}
V\init(do(a,s))\eq y \liff V(time(a), s) \eq y,
\end{align*}
but non-trivially within situations. In the SEA, the four Boolean combinations of $m(s)$ and $w(s)$ constitute the discrete contexts, and the corresponding evolutions are arithmetic combinations of evolutions due to each active process:
\begin{align*}
&V(t,s)\eq y \liff{} \exists v_0. v_0\eq V\init(s) \land \\
&[ \neg m(s)\lland \neg w(s) \lland y\eq v_0\\
& \lor m(s) \land \neg w(s) \land y\eq v_0\!+\!\int_{start(s)}^{t}\hspace{-1em} A(s)\, dt'\\
& \lor \neg m(s) \land w(s) \land y\eq v_0\!-\!\!\int_{start(s)}^{t} \hspace{-2em}\frac{(v(t',s)-50)^2}{10}\, dt'\\
& \lor  m(s) \land w(s) \land y\eq v_0\!+\!\!\int_{start(s)}^{t} \hspace{-2em}A(s) \!-\! \frac{(v(t',s)-50)^2}{10}\, dt'].
\end{align*}
The distance is axiomatized similarly by a SSA for the atemporal fluent $D\init(s)$ and a SEA for the temporal fluent $D(t,s)$.

Since the process $m$ is triggered by the initial state (not shown), every executable situation begins with the natural action $begin_m(0)$. For example, the goal $\exists t(t \geq start(\sigma) \land v(t,\sigma) \geq 10)$ is achieved, among others, by the plan $\sigma \eq do([begin_m(0), accelerate(2)], S_0)$ with $t$ bound to 12.
\end{example}

\section{Discussion}

We proposed an extension of situation calculus which aligns closely with the widely accepted way of representing continuous change in hybrid systems. We then leveraged it to provide a Tarskian semantics to PDDL+. While hybrid automata have a solid foundation in logic \cite{Nerode2007} and notwithstanding the progress toward properly mapping PDDL+ to hybrid automata \cite{bogomolov2014planning}, the PDDL+ semantics, due its focus on grounding, invokes HA in ways which obscure the connection between PDDL+, HA, and logic, putting more emphasis on fine-grained transition system semantics. The lifted semantics proposed in the present paper allows us to sidestep most of the special purpose semantic devices such as happenings and time slippage and remain at a high level of abstraction.

In principle, the proposed hybrid BATs support significantly richer domains than can be specified by PDDL+. For example, SC is bound by neither the closed world nor the domain closure assumption, allowing for domains where objects can be brought in and out of existence. Likewise, hybrid BATs support dynamic systems described by partial DE which are inexpressible by current PDDL+ process effects. At the same time, Tarskian semantics provides for direct verification of various interesting meta-properties of a planning instance, 
 from the essential properties such as satisfiability (due to the relative satisfiability property of hybrid BATs) to testing for mutually exclusive events or establishing a particular relation between executable subsituations. Moreover, a SC axiomatization makes formal causal analysis \cite{causal} applicable to PDDL+ domains.

Notably, PDDL+ does not have a semantics for truly concurrent events, allowing only interleaved concurrency. \cite{reiter} points out examples where interleaved concurrency in SC is not appropriate and develops a theory of true concurrency. In PDDL+, if two events were to occur at the same instant, they would be arbitrarily forced into an order with possibly undesirable consequences. \cite{fl2006} sidestep the problem by considering only \emph{event-deterministic} instances, i.e., those where such nondeterministic choice of order is of no consequence. Although our mapping stays true to the notion of interleaved concurrency for the sake of providing correct semantics, it is equipped to lift this limitation.

Since SC makes all assumptions explicit, it bears one notable syntactic difference to source PDDL+. In particular, the meta-theoretic causal completeness assumption of PDDL+ allows for a very concise description of \emph{relative} process effects. SC affords no such luxury and must list all permutations of all process fluents which may affect a temporal fluent $F$ in the state evolution axiom for $F$. For a worst-case example, consider the road traffic domain of \cite{traffic}. Each road intersection has a set of traffic lights characterized by the process $FlowGreen(r_1, r_2, i)$ parametrized by the relevant road segments $r_1, r_2$ and the intersection $i$. If the road network of a city contains hundreds of intersections, the length of a permutation of all groundings of $FlowGreen$ becomes unmanageable for all practical purposes. However, if one notes (or derives via automated theorem-proving) that only a small fraction of road segments may ever be affected by a particular grounding of $FlowGreen$ (i.e., the maximum number of roads that meet at any one intersection), the SEA is reduced to a manageable size. This appears to be a case of a \emph{sparse predicate} \cite{franco2018} which presents a serious difficulty to PDDL+ planners as well, except it manifests in the explosion of search spaces rather than axioms.

\section{Relevant and Future Work} 

A variety of approaches have been applied to PDDL+ planning. \  \cite{della2009} put forth UPMurphi, an optimal planner based on model checking and blind forward search. They pioneer the discretise-and-validate approach, a method of planning in a discrete approximation of the continuous domain. In case studies \cite{traffic,fox2012plan}, UPMurphi solved challenging problems by relying on domain-specific heuristics.
\cite{BryceGaoMuslinerGoldmanAAAI15}, 
 SMTPlan+ \cite{cashmore2016} and DiNo \cite{piotrowski2016} are the first state-of-the art non-linear planners with full coverage of PDDL+. SMTPlan+ operates by encoding a PDDL+ instance as an SMT problem, inspired by prior successes of non-linear SMT solvers in hybrid systems. Like prior work on PDDL using SAT, this approach reaches state-of-the-art performance by leveraging an existing advanced SMT solver. DiNo builds on and outperforms UPMurphi using a domain-independent heuristic. \cite{balduccini2016} implement a prototype planner using Constraint Answer Set Programming. They achieve noteworthy results, showing that logic programming is a viable approach to PDDL+ planning, especially with domain-specific declarative heuristics. \cite{to2017} introduce MTP, a planner for a new formalism, which is a hybrid of SMT and a metric temporal logic which subsumes PDDL+. Despite allowing goals with rich temporal properties, MTP outperforms SMTPlan+ on standard and new domains. MTP employs a heuristic based on achieving partial goals, which are inexpressible in PDDL+.

All this work points to the fact that PDDL+ planning is significantly more involved than the purely discrete case and might benefit from moving further away from grounding. Recent research focuses on ways of reducing the difficulty by improving modelling techniques \cite{denenberg2018} or automatically reducing the arity of predicates with large numbers of possible groundings \cite{franco2018}. Although generally less tractable than the PDDL family of languages, situation calculus has been used successfully for planning purposes in \cite{wspdf}. There, the focus of the \ttt{wspdf} planner was to exploit the expressiveness of SC to extend practical planning to open worlds. 
We posit that native SC planning based on theorem proving holds promise for PDDL+ domains as well. 
We imagine that an effective SC planner can be constructed by supplementing \ttt{wspdf} with 
a non-linear optimizer such as IPopt \cite{ipopt}, resulting in an iteratively deepening forward search interleaved with solving initial value problems.  Tractability of theorem proving can be achieved by considering \emph{bounded} hybrid BATs along the lines of \cite{DeGiacomoLesperancePatriziAIJ16}. The present paper can be considered as a step in this direction.

\bibliography{literature.bib}
\bibliographystyle{plain}
\end{document}